\documentclass[sigconf]{acmart}

\copyrightyear{2026}
\acmYear{2026}
\setcopyright{cc}
\setcctype{by}
\acmConference[MMSports '26]{9th Int. Workshop on Multimedia Content Analysis in Sports}{November 10--14, 2026}{Rio de Janeiro, Brazil}
\acmBooktitle{9th Int. Workshop on Multimedia Content Analysis in Sports (MMSports '26), November 10--14, 2026, Rio de Janeiro, Brazil}
\acmDOI{10.1145/3841455.3841525}
\acmISBN{979-8-4007-2944-7/2026/11}

\AtBeginDocument{%
  }

\usepackage{booktabs}
\usepackage{multirow}
\usepackage{float}

\begin{document}

\title{BMASH: Ball-Motion-Aware Soccer Header Spotting}

\author{Ahmed Endris Hasen}
\authornote{Corresponding author.}
\email{ahmed.e.hasen@jyu.fi}
\affiliation{%
  \institution{Faculty of Information Technology,\\ University of Jyväskylä}
  \city{Jyväskylä}
  \country{Finland}
}

\author{Muhammad Shahzad Khan}
\email{muhammad.s.khan@jyu.fi}
\affiliation{%
  \institution{Faculty of Information Technology,\\ University of Jyväskylä}
  \city{Jyväskylä}
  \country{Finland}
}

\author{Nikolaos Passalis}
\email{passalis@csd.auth.gr}
\affiliation{%
  \institution{Faculty of Sciences,\\ Aristotle University of Thessaloniki}
  \city{Thessaloniki}
  \country{Greece}
}

\author{Jenni Raitoharju}
\email{jenni.k.raitoharju@jyu.fi}
\affiliation{%
  \institution{Faculty of Information Technology,\\ University of Jyväskylä}
  \city{Jyväskylä}
  \country{Finland}
}

\renewcommand{\shortauthors}{Hasen et al.}

\begin{abstract}
Recent advances in computer vision have made broadcast sports videos increasingly useful for event analysis, performance assessment, and player-safety applications. In soccer, however, header spotting remains a challenging problem due to the subtle and short-lived nature of header events. This paper focuses on soccer header spotting: identifying moments in broadcast videos where the ball contacts a player's head. We first adapt and evaluate Video Swin as a strong action-recognition baseline for this task, and then introduce BMASH, a ball-motion-aware fusion framework that integrates detector-derived ball features. BMASH combines Video Swin action representations with ball-presence and motion features from frame-level soccer-ball detections, integrating player-action context with ball dynamics to distinguish headers from visually similar events. We evaluate BMASH using game-level splits with separate test matches and rotating validation folds, considering both centered-window classification and continuous full-video spotting. Results show that Video Swin provides a strong baseline for header spotting, while BMASH improves clip-level AP and ROC-AUC over the corresponding Video Swin baseline. In continuous full-video spotting, BMASH achieves a comparable event-level F1-performance with a different precision--recall trade-off.
\end{abstract}




\begin{teaserfigure}
  \includegraphics[width=\textwidth]{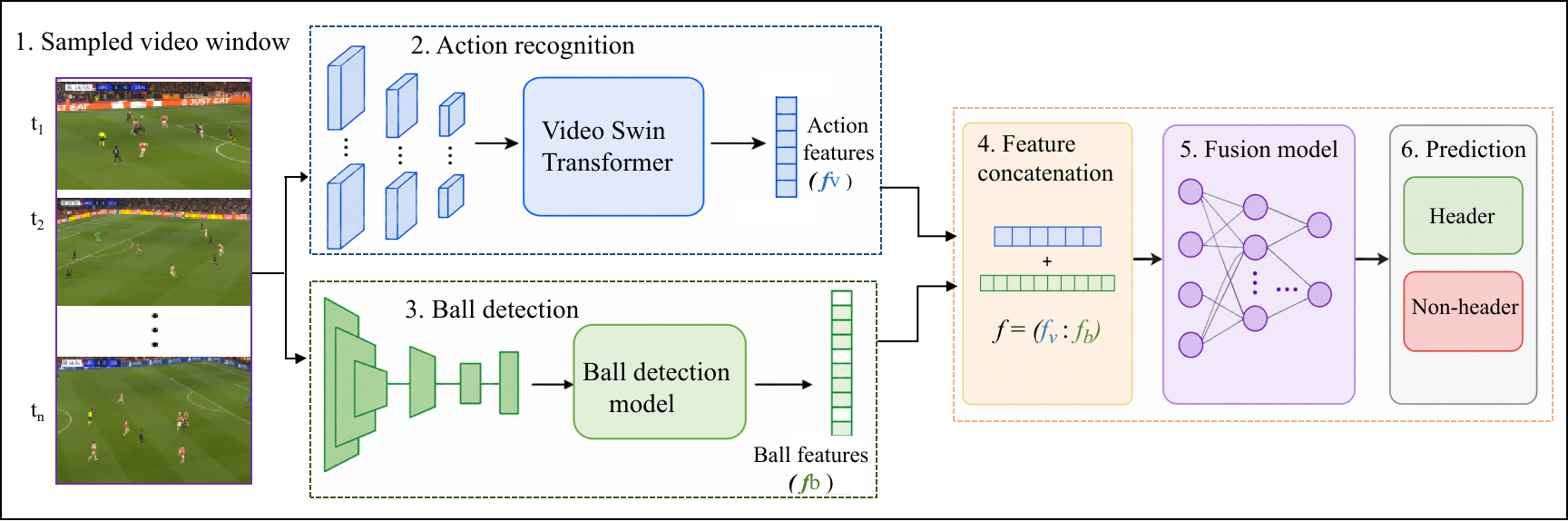}
   \caption{Overview of the proposed BMASH framework. The action-recognition branch and the ball-detection branch are trained separately using task-specific data. Given a sampled video window, the Video Swin branch processes the full stack of $T$ frames and produces action-score features $\mathbf{f}_v$, while the YOLO ball detector is applied frame-wise and its detections are summarized into compact ball-motion features $\mathbf{f}_b$. The two feature vectors are concatenated and passed to a lightweight fusion MLP to predict whether the input window contains a header.}
  \Description{BMASH contains a Video Swin branch for action features and a ball-detection branch for ball features. Their outputs are fused to predict header or non-header. }
  \label{fig:bmash_overview}
\end{teaserfigure}

\maketitle

\section{Introduction}

Recently, progress in sports video understanding has been driven by large-scale benchmarks, advances in spatiotemporal deep learning, and automated broadcast analysis~\cite{naik2022sportscv, akan2023use}. These developments have substantially expanded the scope of computer vision in sports, enabling applications such as event recognition, automatic highlight generation, tactical analysis, player tracking, performance assessment, and athlete health monitoring~\cite{zheng2025review, xu2025deep}. Among these applications, action spotting has become a fundamental problem in soccer video understanding, aiming to localize the occurrence of semantic events within broadcast matches~\cite{xu2025deep,giancola2018soccernet, deliege2021soccernet,giancola2025deep}. Large-scale benchmarks, such as SoccerNet~\cite{giancola2018soccernet}, have accelerated research in this direction by providing annotated broadcast matches for evaluating event localization methods.

Despite this progress, soccer header spotting remains considerably less explored than other action spotting. A soccer header is characterized by a brief ball-head interaction that frequently lasts only a few frames and occupies a very small image region~\cite{rezaei2022automated}. Detecting this subtle interaction, therefore, represents a challenging fine-grained recognition problem that requires reasoning about player motion, surrounding context, and ball dynamics rather than relying solely on global scene appearance~\cite{giancola2023towards,xu2024finesports}.

The difficulty is further affected by the characteristics of broadcast soccer videos. Camera motion, rapid zooming, viewpoint changes, motion blur, occlusion, compression artifacts, and limited spatial resolution of distant players all reduce the visibility of both the ball and the head-ball contact~\cite{giancola2018soccernet}. Moreover, many visually similar actions, including crosses, clearances, long passes, aerial duels, throw-ins, and shots, share highly similar motion patterns and ball trajectories, making it difficult to distinguish heading events from other airborne ball interactions~\cite{giancola2025deep,rezaei2022automated}. 

Beyond its importance for automatic match analysis, reliable header spotting is increasingly motivated by player-safety research. Numerous biomechanical and epidemiological studies have investigated repetitive soccer headers and their potential relationship with head-impact exposure, concussion risk, and long-term neurological outcomes~\cite{kenny2022headimpactvarsity}. However, estimating heading exposure currently relies heavily on labor-intensive manual video review or wearable sensing followed by extensive human verification~\cite{kenny2022headimpactvarsity, kern2022neural}. Recent computer vision systems, such as DeepImpact~\cite{rezaei2022automated}, have demonstrated that automated video screening can substantially reduce this manual effort by identifying candidate heading events from broadcast footage. Consequently, accurate header spotting constitutes an essential first stage toward scalable video-based head-impact monitoring and downstream biomechanical analysis.

Despite recent progress in sports video understanding, soccer header spotting remains comparatively underexplored as a standalone fine-grained recognition problem. Reliable soccer header spotting requires both accurate spatiotemporal understanding of player actions and awareness of the ball's presence and movement. Modern video recognition architectures can capture player motion, body interactions, and broadcast context, while ball detections provide additional information about the ball's presence and movement. However, ball information alone is not sufficient, since many non-header actions also involve airborne ball motion. Thus, ball motion is used as supporting evidence for visual action features.

To address these challenges, we adapt Video Swin to the soccer header spotting task and introduce \textbf{BMASH} (\textbf{B}all-\textbf{M}otion-\textbf{A}ware \textbf{S}occer \textbf{H}eader Spotting), a lightweight late-fusion framework that integrates full-frame action representations with detector-derived ball-motion information. BMASH combines spatiotemporal Video Swin action representations with ball-motion information derived from frame-level soccer-ball detections.

We evaluate BMASH on a soccer header dataset constructed from multiple broadcast matches. To ensure a realistic evaluation and avoid overlap between training and test footage, we use game-level splits with unseen test matches. Evaluation is conducted at both the temporal-window level, using labeled header and non-header clips, and the full-video level, where the model is applied to continuous broadcast video using sliding-window inference.
The main contributions of this work are:
\begin{itemize}
   \item We adapt Video Swin as a strong transformer-based action-recognition baseline for soccer header spotting and compare it with EfficientNet3D and TSM-ResNet50 under a unified header/non-header protocol.

   \item We propose \textbf{BMASH}, a lightweight ball-motion-aware late-fusion framework that combines Video Swin-based action representations with compact detector-derived ball features for soccer header spotting.

   \item We evaluate the models using game-level splits with unseen test videos across different input-window configurations, and further analyze their performance in continuous full-video spotting.
\end{itemize}

The rest of the paper is organized as follows. Section~\ref{sec:related_work} discusses prior and related works on action spotting, header detection, and head-impact analysis. Section~\ref{sec:method} presents the BMASH framework. Section~\ref{sec:experiments} describes the dataset, compared methods, and experimental setup. Section~\ref{sec:results} presents the results and analysis. Section~\ref{sec:conclusion} concludes the paper and discusses future work.

\section{Related Work}
\label{sec:related_work}
\subsection{Action Spotting in Soccer}

Action spotting aims to temporally localize semantic events within long, untrimmed sports videos. In soccer, this problem has become one of the core tasks in video understanding following the introduction of SoccerNet~\cite{giancola2018soccernet}, which established the first large-scale benchmark for temporal localization of events in full broadcast matches. SoccerNet-v2~\cite{deliege2021soccernet} further expanded both the annotation scale and benchmark tasks, enabling broader research on long-form soccer video understanding. Recent surveys show that SoccerNet is the most widely used benchmark for soccer action spotting and temporal event localization~\cite{giancola2025deep,zheng2025review}.

Research has progressively shifted from learning global clip representations toward explicitly modeling temporal context surrounding candidate events. NetVLAD++~\cite{giancola2021temporally} introduced temporally-aware feature aggregation by separately encoding information before and after candidate timestamps. Cioppa et al.~\cite{cioppa2020context} introduced a context-aware loss function (CALF) for soccer action spotting, which models the temporal neighborhood around annotated events rather than treating each timestamp in isolation. Subsequent methods explored multiple-scene representations to exploit broadcast transitions~\cite{shi2022action}, dense temporal anchor prediction for precise localization~\cite{soares2022temporally}, and transformer-based reasoning capable of modeling long-range temporal dependencies while addressing class imbalance and temporal uncertainty~\cite{xarles2023astra}. 
Unified frameworks, such as OSL-ActionSpotting~\cite{benzakour2024osl}, have further improved soccer action spotting by providing standardized implementations of major spotting algorithms.

Recent SoccerNet challenges have shifted attention toward denser ball-centric understanding by introducing Ball Action Spotting, where headers constitute only one of several fine-grained ball interaction classes alongside passes, shots, throw-ins, goals and others~\cite{cioppa2024soccernet}. Cioppa et al.~\cite{cioppa2024soccernet} presented the best-performing EfficientNet-based 2D--3D model, where 2D convolutions encode appearance from individual frames and 3D convolutions model temporal information across stacked frames. Parallel research has also investigated anticipating future ball actions before they occur~\cite{dalal2025action}. 

These studies demonstrate growing interest in modeling subtle ball-player interactions. However, most existing approaches target large-scale, multi-class event spotting, where headers constitute only one of many event categories and can be difficult to model due to class imbalance and limited examples. In contrast, our work focuses on dedicated binary soccer header spotting as a standalone fine-grained recognition problem.

\subsection{Soccer Header and Head-Impact Detection}
Compared with general soccer action spotting, dedicated computer vision methods for automated soccer header detection have received considerably less attention, leaving a gap between soccer action spotting and automated head-impact analysis. Prior work has explored computer vision pipelines for estimating soccer head-impact exposure from broadcast video, including approaches that combine ball detection, tracking, localized video crops, and temporal classification to identify potential heading events~\cite{rezaei2022automated}. Although the approach achieved high header detection sensitivity on full-match tests, the reported precision dropped to  21.1\% due to a large number of false-positive detections, highlighting the difficulty of robust header spotting in real full match scenarios~\cite{rezaei2022automated}. In such pipelines, the temporal classifier receives cropped video regions centered on the detected ball, so missed or incorrect ball detections can directly lead to crops that omit the relevant header action or focus on an irrelevant region. These studies show the feasibility of using broadcast footage for header-related analysis, while also motivating further work on robust header spotting in realistic match videos. In contrast to ball-centered crop pipelines, our framework preserves full-frame spatiotemporal reasoning and uses detector-derived ball information as supporting evidence through late fusion.

Beyond computer vision, extensive biomechanics research has investigated soccer heading through wearable sensing. Kern et al.~\cite{kern2022neural} developed a neural network capable of distinguishing true headers from false-positive recordings from xPatch wearable head-impact sensors placed behind the players' right ear over the mastoid process, using kinematic measurements verified by broadcast video. Kenny et al.~\cite{kenny2022headimpactvarsity} quantified heading biomechanics in collegiate soccer using an instrumented mouthpiece with tri-axis accelerometers and gyroscopes, synchronized with video verification. In a subsequent longitudinal study, the same group investigated individual heading exposure using a large set of video-confirmed headers collected over an extended observation period~\cite{kenny2024individualized}.
Additional investigations have characterized heading burden and head-injury-risk situations in elite football and extended exposure analysis to professional football~\cite{filben2021header}. The broader head-impact literature consistently demonstrates that wearable sensors alone generate substantial numbers of false-positive events, making video confirmation essential for reliable exposure estimation~\cite{kern2022neural, le2022head}. More recently, quantitative video analysis has been recognized as an effective complementary alternative for scalable head-impact monitoring without wearable sensors and extensive manual annotation~\cite{aston2025quantitative}. 

\subsection{Video Action Recognition and Ball Tracking}

Deep learning for video understanding has evolved from convolution-based architectures toward transformer-based spatiotemporal representation learning. Early approaches, such as the Temporal Shift Module (TSM)~\cite{lin2019tsm}, introduced temporal reasoning into 2D convolutional networks by shifting feature channels across adjacent frames. More recent transformer-based architectures have substantially improved spatiotemporal video understanding. Among these, Video Swin Transformer~\cite{liu2022video} employs hierarchical shifted-window attention to efficiently model long-range spatiotemporal patterns, making it a strong backbone for dedicated soccer header spotting.

Soccer ball detection and tracking have also been widely studied as the ball is small, fast-moving, frequently blurred, and often occluded in broadcast footage~\cite{komorowski2019deepball,kamble2019ball, kamble2019deep}. DeepBall~\cite{komorowski2019deepball} introduced a dedicated deep-learning framework for long-shot soccer ball detection, while subsequent work used temporal consistency to improve robustness when the ball is temporarily invisible~\cite{kamble2019ball}. Semi-supervised learning has further improved player and ball detection in SoccerNet broadcasts~\cite{vandeghen2022semi}.

Prior work in other ball sports also highlights the value of temporal object modeling. TrackNet~\cite{huang2019tracknet} showed that temporal information improves localization of small, high-speed tennis balls in broadcast video, while TrackNetV4~\cite{raj2025tracknetv4} further explored motion-aware feature fusion for ball tracking in tennis, badminton, and table tennis. These studies provide useful motivation for incorporating ball-motion information into header spotting, and our work investigates whether compact detector-derived ball-motion descriptors can complement transformer-based video representations for dedicated soccer header spotting.

\section{Method}
\label{sec:method}

\subsection{Overview of our Method}
\label{sec:method_overview}

BMASH is a ball-motion-aware late-fusion framework for binary soccer header spotting in broadcast video. Given a short temporal window sampled from a match, the model predicts whether the window contains a header or a visually similar non-header action. The task is challenging because the defining action evidence, namely ball-head contact, is often brief, small in the image, and partially obscured by players, camera motion, or motion blur. At the same time, many non-header actions also exhibit similar airborne ball motion, making ball visibility alone insufficient for reliable prediction. BMASH then combines full-frame action understanding with detector-derived ball information as supporting evidence.

We denote the sampled window as $\mathbf{X}=\{\mathbf{x}_{t_1},\mathbf{x}_{t_2},\ldots,\mathbf{x}_{t_T}\}$ and assign it a binary label $y \in \{0,1\}$, where $y=1$ denotes a header and $y=0$ denotes a non-header action. The goal is to learn a classifier $F(\mathbf{X})$ that estimates
\begin{equation}
    \hat{y} = F(\mathbf{X}) = P(y=1 \mid \mathbf{X}),
\end{equation}
where $\hat{y}$ is the predicted probability that the input window contains a header.

To address this, BMASH combines two complementary sources of information from the same input window. The first source is a global spatiotemporal action representation extracted from the full broadcast clip using a Video Swin Transformer. This branch captures player motion, body configuration, player-player interaction, and surrounding scene context. The second source is an explicit ball-motion representation derived from frame-level soccer-ball detections, summarizing ball visibility and movement over time. These window-level representations are aligned and fused through a lightweight classifier, allowing explicit ball information to complement full-frame visual action understanding.

Figure~\ref{fig:bmash_overview} illustrates the overall BMASH pipeline. A video window is sampled into $T$ frames and processed by both branches. The Video Swin branch produces a 6-dimensional action-score feature vector, while the ball-detection branch summarizes detector outputs into a 10-dimensional ball-motion descriptor. The two representations are concatenated and passed through a lightweight fusion MLP that predicts the probability that the input window contains a header. This late-fusion design allows detector-derived ball information to complement full-frame spatiotemporal action understanding without making the final prediction depend only on the presence of a detected ball.

\subsection{Video Swin Action Recognition}
\label{sec:visual_branch}

The Video Swin branch provides the  spatiotemporal action representation used by BMASH. It learns window-level video representations from the sampled frames.  Given an input video window, we sample $T$ frames and denote the resulting clip as
$\mathbf{X}=\{\mathbf{x}_{t_1}, \mathbf{x}_{t_2}, \ldots, \mathbf{x}_{t_T}\}$,
where $\mathbf{x}_{t_i}$ is the frame sampled at time $t_i$. The sampled frames are processed together as a spatiotemporal clip.

Video Swin processes the complete sampled window, allowing the model to capture player motion, body pose changes, nearby player interactions, and broader scene context. The Video Swin encoder processes $\mathbf{X}$ and outputs two-class softmax scores, denoted by $p_{NH}$ and $p_H$ for the non-header and header classes. When multiple temporal clips are sampled from the same input window, the same notation refers to the scores from each sampled clip. These scores are complementary probabilities. For fusion, we summarize them into a 6-dimensional score feature vector:
\begin{equation}
\mathbf{f}_v =
[\mu_{NH}, \mu_H, m_{NH}, m_H, \sigma_{NH}, \sigma_H],
\end{equation}
% $\mathbf{f}_v=[\mu_{NH},\mu_H,m_{NH},m_H,\sigma_{NH},\sigma_H]$,
where $\mu$, $m$, and $\sigma$ denote the mean, maximum, and standard deviation of the non-header and header scores across the temporal clips sampled from the same window. Thus, $\mathbf{f}_v \in \mathbb{R}^{6}$.

Using the sampled video window is important because soccer headers are defined not only by the ball location, but also by the surrounding action context. Body orientation, aerial duels, nearby opponents, and the phase of play can help distinguish headers from other ball-in-play actions. However, the actual ball-head contact can still be small and brief, so BMASH complements the action representation with an explicit ball-motion branch.

\subsection{Ball Detection and Motion Encoding}
\label{sec:ball_branch}
The ball-detection and motion descriptor branch provides explicit ball-related information from the same sampled video window. We use a YOLO11s~\cite{yolo11_ultralytics} object detector pretrained on the COCO dataset~\cite{lin2014microsoft} and fine-tuned for soccer-ball detection using separate SoccerNet tracking data. The fine-tuned ball detector is then applied to each sampled frame in the input window to detect the ball. For each frame, we keep the highest-confidence ball candidate and encode its detection information, i.e., the detection confidence, normalized position and normalized bounding-box area. If no ball candidate is detected, the detection values are set to zero. The frames are processed independently without any additional clip-level adjustment of the detections.

For each sampled frame $t_i$, the detector output is represented as
$\mathbf{b}_{t_i}=[r_i,c_i,x_i,y_i,a_i]$, where $r_i$ indicates whether the ball is detected, $c_i$ is the detection confidence, $(x_i,y_i)$ is the normalized ball-center position, and $a_i$ is the normalized bounding-box area. If no ball is detected, all values are set to zero. Across the input window, the frame-level detections
$\{\mathbf{b}_{t_1}, \mathbf{b}_{t_2}, \ldots, \mathbf{b}_{t_T}\}$
are summarized into a 10-dimensional ball-motion descriptor:
\begin{equation}
\mathbf{f}_b =
[\mu_c, m_c, \sigma_c, \mu_r, \mu_a, m_a, \mu_x, \mu_y, \mu_d, m_d],
\end{equation}
where $\mu_c$, $m_c$, and $\sigma_c$ are the mean, maximum, and standard deviation of the ball-detection confidence; $\mu_r$ is the detection rate; $\mu_a$ and $m_a$ are the mean and maximum normalized bounding-box area; $\mu_x$ and $\mu_y$ are the mean normalized ball-center positions; and $\mu_d$ and $m_d$ are the mean and maximum displacement between consecutive detected ball centers. Thus, $\mathbf{f}_b \in \mathbb{R}^{10}$.

Ball detections are used as supporting evidence rather than as a standalone decision signal. This is important because visible ball motion appears in many non-header events, while true headers may still have weak or missing ball detections due to occlusion, blur, or small ball size. The ball-detection branch, thus, provides structured ball-related information that complements the Video Swin based action representation learned from the sampled video window.

\subsection{Ball-Motion-Aware Fusion}
\label{sec:fusion_classifier}

BMASH combines the two window-level feature vectors using late fusion. Since the Video Swin branch and the ball-motion branch operate on the same sampled video window $\mathbf{X}$, their outputs are aligned at the window level.  The final fusion input is obtained by concatenating the 6-dimensional Video Swin action-score feature vector $\mathbf{f}_v$ and the 10-dimensional ball-motion descriptor $\mathbf{f}_b$:
\begin{equation}
    \mathbf{f} = [\mathbf{f}_v ; \mathbf{f}_b] \in \mathbb{R}^{16},
\end{equation}
where $[\cdot ; \cdot]$ denotes feature concatenation. The fused 16-dimensional vector is passed to a lightweight multi-layer perceptron (MLP) classifier:
\begin{equation}
    \hat{y} = C(\mathbf{f}),
\end{equation}
where $\hat{y}$ is the predicted probability that the input window contains a header and $C(\cdot)$ denotes the fusion MLP classifier. The MLP consists of two hidden layers with ReLU activations and dropout, followed by a single output unit whose sigmoid activation gives the header probability. During fusion training, the Video Swin and ball-detection branches were kept fixed, and only the fusion MLP was optimized using the extracted feature representations.

The fusion classifier is trained using weighted binary cross-entropy with logits. We use class weighting to account for differences in the number of header and non-header samples within each training fold. The header-class weight $w_h$ is computed as the ratio of non-header to header samples in the training fold.  The loss can be written as
\begin{equation}
    \mathcal{L} =
    - w_h y \log(\hat{y})
    - (1-y)\log(1-\hat{y}),
\end{equation}
where $y \in \{0,1\}$ is the ground-truth label and $\hat{y}$ is the predicted header probability. When threshold tuning is used, the decision threshold is selected on the validation split and then applied to the held-out test set.

The fusion classifier learns to combine the two evidence sources. Strong ball motion without compatible player posture may correspond to a cross, clearance, long pass, or shot. Conversely, a visually plausible header may still occur when the ball detector is uncertain due to blur, occlusion, or small ball size. The late-fusion design allows BMASH to use ball-motion information when it supports the action context, without treating ball detections as a standalone decision rule. When ball evidence is weak or missing, the classifier can rely more on the Video Swin action representation. 

\section{Experiments}
\label{sec:experiments}

\subsection{Dataset}
\label{sec:dataset}

The dataset was constructed from SoccerNet broadcast videos recorded at 25 fps. We used all the seven matches from the SoccerNet Ball Action Spotting annotations, from which header timestamps were extracted. To increase data diversity, we additionally added five raw SoccerNet broadcast matches (Wigan Athletic--Birmingham City, Cardiff City--Queens Park Rangers, Napoli--Inter Milan, Real Madrid--1. FC Union Berlin, Arsenal--Racing Club de Lens)   and manually annotated header timestamps. The selected matches were split at the game level, so that windows from the same match do not appear in both training and test sets. 

Each annotated timestamp was converted into a short temporal window. Header windows were centered on visible ball-head contact. Candidate events were reviewed frame by frame, and unclear cases were excluded when ball-head contact could not be verified because of severe occlusion or camera cuts. Non-header windows were sampled from visually similar ball-in-play situations, such as crosses, clearances, passes, shots, throw-ins, aerial duels, and other airborne-ball interactions, while ensuring that the selected windows did not contain visible ball-head contact. This centered-window setting isolates the recognition problem by evaluating whether a candidate temporal window contains a header or a visually similar non-header action. The separate full-video analysis in Section~\ref{sec:full_video_results} evaluates the complete spotting setting, where candidate windows are generated from continuous broadcast video using sliding-window inference.

The test set remained fixed across all folds and contains two matches: BN (Blackburn Rovers--Nottingham Forest) and WB (Wigan Athletic--Birmingham City), a total of 529 temporal windows derived from annotated timestamps, with 261 headers and 268 non-headers. The remaining ten matches formed the training/validation pool. For rotating game-level validation, each fold used one match for validation and the other nine matches for training. Table~\ref{tab:dataset_summary} summarizes the class distribution across the training/validation pool and the fixed test set, while Table~\ref{tab:validation_folds} lists the validation match used in each fold.

\begin{table}[t]
\centering
\caption{Distribution of the dataset across the experiment.}
\label{tab:dataset_summary}
\small
\begin{tabular}{lccc}
\toprule
Split & Non-header & Header & Total \\
\midrule
Train/validation pool & 1067 & 947 & 2014 \\
Test set & 268 & 261 & 529 \\
\midrule
Total & 1335 & 1208 & 2543 \\
\bottomrule
\end{tabular}
\end{table}

\begin{table}[t]
\centering
\caption{Rotating game-level validation folds. Counts are given as header/non-header.}
\label{tab:validation_folds}
\small
\resizebox{\columnwidth}{!}{
\begin{tabular}{clcc}
\toprule
Fold & Validation game & Train & Validation \\
\midrule
0 & Brentford--Bristol City & 847/957 & 100/110 \\
1 & Hull City--Sheffield Wednesday & 822/957 & 125/110 \\
2 & Leeds United--West Bromwich & 879/957 & 68/110 \\
3 & Middlesbrough--Preston North End & 820/957 & 127/110 \\
4 & Reading--Fulham & 893/957 & 54/110 \\
5 & Stoke City--Huddersfield Town & 819/957 & 128/110 \\
6 & Cardiff City--Queens Park Rangers & 839/949 & 108/118 \\
7 & Napoli--Inter Milan & 856/959 & 91/108 \\
8 & Real Madrid--1. FC Union Berlin & 861/965 & 86/102 \\
9 & Arsenal--Racing Club de Lens & 887/988 & 60/79 \\
\bottomrule
\end{tabular}
}
\end{table}

\subsection{Compared Methods}
\label{sec:comparisons}

While many methods have been proposed for soccer action spotting, our comparison focuses on models that are most relevant to header spotting: ball-action spotting models that include header as one of the action classes, a header specific recognition method, and Video Swin as a main action-recognition baseline.

For generic ball-action spotting, we evaluate a 12-class SoccerNet Ball Action EfficientNet3D model~\cite{cioppa2024soccernet}, where header is one of several classes, including pass, drive, high pass, cross, shot, and throw-in. In this Ball Action Spotting setting, header obtains approximately 0.56 AP@1, indicating that headers are among the more difficult ball-action classes in full-video event spotting. On our centered clip-level test set, the same header-score evaluation gives high precision but low recall at the default 0.5 threshold (precision = 0.9230, recall = 0.4303, F1 = 0.5870). This comparison addresses whether a specialized binary header detector provides an advantage over a generic ball-action model in which header is only one class among several actions. Similarly, prior header-detection work using TSM-ResNet50 with ball tracking and ball-centered cropping reported high sensitivity but only 21.1\% precision, highlighting the difficulty of realistic full-video header detection~\cite{rezaei2022automated}.

For direct comparison on our dataset, we trained and evaluated EfficientNet3D, TSM-ResNet50, and Video Swin~\cite{liu2022video} under the same game-level binary header and non-header protocol. EfficientNet3D provides a convolutional spatiotemporal baseline, TSM-ResNet50 represents a temporal-shift 2D CNN baseline related to prior header-detection work, and Video Swin serves as the strongest action-recognition baseline. BMASH uses the Video Swin branch and adds detector-derived ball-motion features through late fusion, allowing us to compare full-frame action recognition with ball-motion-aware fusion under the same dataset, split, and evaluation protocol.

\subsection{Experimental Setup}
\label{sec:exp_setup}
This section presents the experimental setup and implementation details for BMASH and the baseline models. All models were trained on temporal windows from the training matches, selected using rotating validation folds, and tested on the same fixed held-out matches. For each fold, one match was used for validation, and the checkpoint with the best validation performance was used as the final model for test evaluation. This procedure produced ten independently trained models, one for each rotating validation match. Table~\ref{tab:main_clip_results} reports the mean and standard deviation of the fixed test-set metrics across these ten trained models. 
EfficientNet3D and TSM-ResNet50 used 33-frame and 17-frame windows, respectively. Video Swin (along with BMASH) used two temporal settings, with 16-frame and 32-frame windows, to analyze the effect of input-window length.

EfficientNet3D and TSM-ResNet50 were trained using AdamW~\cite{loshchilov2017decoupled} for 30 epochs with learning rates of $1\times10^{-3}$ and $3\times10^{-4}$, respectively. Video Swin~\cite{liu2022video} was implemented based on MMAction2~\cite{2020mmaction2} using a Video Swin-Tiny backbone pretrained on Kinetics-400 and fine-tuned on our soccer header dataset for header spotting task. The model was trained for 30 epochs using AdamW with a learning rate of $1\times10^{-3}$. Video Swin-16 uses 16-frame clips at $960{\times}544$, while Video Swin-32 uses 32-frame clips at $1280{\times}736$ with gradient accumulation. For BMASH, the fusion MLP was trained for 100 epochs using the Video Swin action and ball-motion features, binary cross-entropy with logits, AdamW, and batch size 16. The ball detector used a YOLO11s~\cite{yolo11_ultralytics}  model pretrained on COCO~\cite{lin2014microsoft} and fine-tuned for soccer-ball detection for 20 epochs with input size 1280, using batch size 6 and learning rate $0.01$.

All experiments were conducted on the CSC Roihu environment using NVIDIA GH200 GPU nodes with Python 3.12, PyTorch 2.10, MMAction2/MMEngine, and Ultralytics YOLO.

\subsection{Evaluation Metrics}
\label{sec:metrics}

We evaluated the performance using average precision (AP), F1-score, precision, recall, balanced accuracy, and ROC-AUC. AP and ROC-AUC summarize how well the model ranks header and non-header windows independently of a single decision threshold. Precision, recall, F1-score, and balanced accuracy measure classification performance after converting model scores into binary predictions. 

Full-video spotting was evaluated separately as an event-detection task. The model was applied to continuous broadcast videos using sliding-window inference with 2.0-second windows and a stride of 0.5 seconds. Window-level scores were  smoothed and processed with non-maximum suppression to merge nearby detections and prevent repeated predictions for the same event. Predicted events were matched with annotated header timestamps using a tolerance of $\pm 2$ seconds. Matched predictions were counted as true positives, unmatched predictions as false positives, and unmatched annotations as false negatives. We report event-level precision, recall, F1-score, and false positives per match. True negatives were not used for event-level full-video evaluation, since continuous match footage contains a very large number of non-header moments.

\section{Results Analysis}
\label{sec:results}

\subsection{Overall Performance}
\label{sec:main_results}

Table~\ref{tab:main_clip_results} shows the  performance on the test set. Results are reported as mean $\pm$ standard deviation across the validation configurations reported in Table~\ref{tab:validation_folds}.
TSM-ResNet50 achieves high recall but low precision and balanced accuracy, indicating that it tends to over-predict the header class. Video Swin provides a much stronger visual action baseline, improving AP and ROC-AUC. Adding ball-derived features produces a clear improvement in the 16-frame setting: BMASH-16 increases AP from 0.8858 to 0.8966 and F1-score from 0.7416 to 0.7961. The 32-frame configuration produces a stronger visual baseline, with Video Swin-32 reaching 0.9423 AP and 0.8636 F1-score. BMASH-32 achieves the highest AP and ROC-AUC overall, reaching 0.9530 AP and 0.9507 ROC-AUC, while maintaining essentially the same F1-score as Video Swin-32. These results indicate that detector-derived ball information primarily improves score ranking and class separation, whereas the strongest visual backbone already provides highly competitive threshold-based classification performance.

\begin{table*}[t]
\centering
\caption{Performance comparison of all evaluated models. Results are reported as mean $\pm$ standard deviation across the validation folds. The best result for each metric is shown in \textbf{bold}.}
\label{tab:main_clip_results}
\small
\resizebox{\textwidth}{!}{
\begin{tabular}{lcccccc}
\toprule
\textbf{Method} & \textbf{AP} & \textbf{F1} & \textbf{Precision}  & \textbf{Recall} & \textbf{Balanced Acc.} & \textbf{ROC-AUC}\\

\midrule
EfficientNet3D & $0.7061{\pm}0.0544$ & $0.6606{\pm}0.0429$ & $0.5927{\pm}0.0699$ & $0.7931{\pm}0.1809$ & $0.6094{\pm}0.0517$ & $0.7124{\pm}0.0517$ \\
TSM-ResNet50 & $0.5889{\pm}0.0642$ & $0.6631{\pm}0.0034$ & $0.5008{\pm}0.0083$ & $\mathbf{0.9820{\pm}0.0206}$ & $0.5138{\pm}0.0153$ & $0.5913{\pm}0.0457$ \\
Video Swin-16 & $0.8858{\pm}0.0250$ & $0.7416{\pm}0.0683$ & $0.8786{\pm}0.0463$ & $0.6525{\pm}0.1040$ & $0.7794{\pm}0.0388$ & $0.8831{\pm}0.0247$ \\
BMASH-16 & $0.8966{\pm}0.0244$ & $0.7961{\pm}0.0575$ & $0.8096{\pm}0.0691$ & $0.8023{\pm}0.1177$ & $0.8013{\pm}0.0416$ & $0.8990{\pm}0.0223$ \\
Video Swin-32 & $0.9423{\pm}0.0131$ & $0.8636{\pm}0.0364$ & $0.9022{\pm}0.0280$ & $0.8333{\pm}0.0749$ & $0.8715{\pm}0.0279$ & $0.9418{\pm}0.0139$ \\
\textbf{BMASH-32} & $\mathbf{0.9530{\pm}0.0103}$ & $\mathbf{0.8639{\pm}0.0383}$ & $\mathbf{0.9152{\pm}0.0290}$ & $0.8241{\pm}0.0812$ & $\mathbf{0.8735{\pm}0.0289}$ & $\mathbf{0.9507{\pm}0.0114}$ \\
\bottomrule
\end{tabular}
}
\end{table*}

\subsection{Per-Fold and Qualitative Analysis}
\label{sec:fold_full_video}

\begin{figure}[t]
\centering
\includegraphics[width=\linewidth]{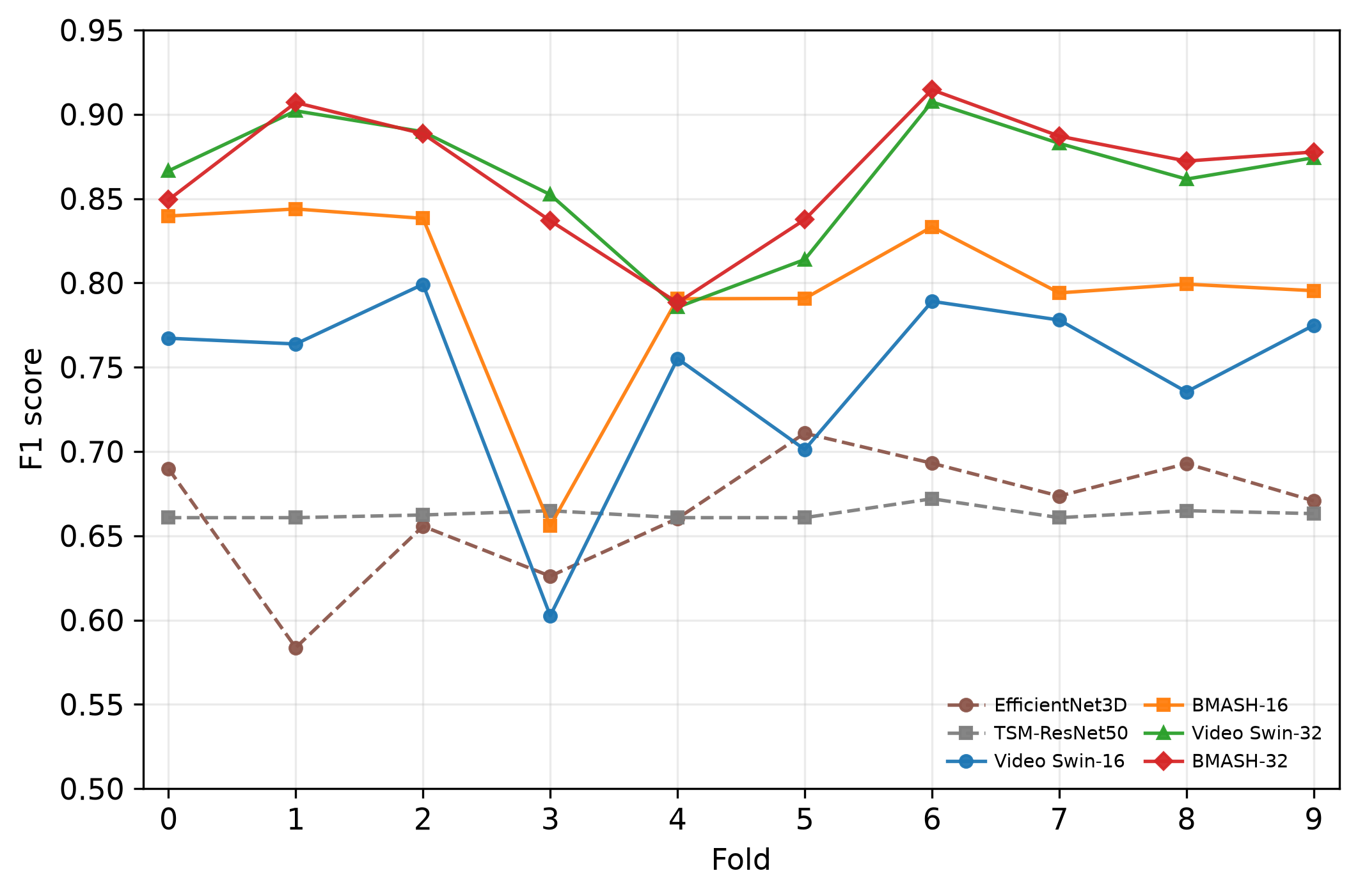}
\caption{Per-fold F1-score comparison of the proposed BMASH models and the baseline methods.}
\Description{F1 scores across evaluation folds for the BMASH models and baseline methods}
\label{fig:per_fold_f1}
\end{figure}

\begin{figure}[t]
\centering
\includegraphics[width=\linewidth]{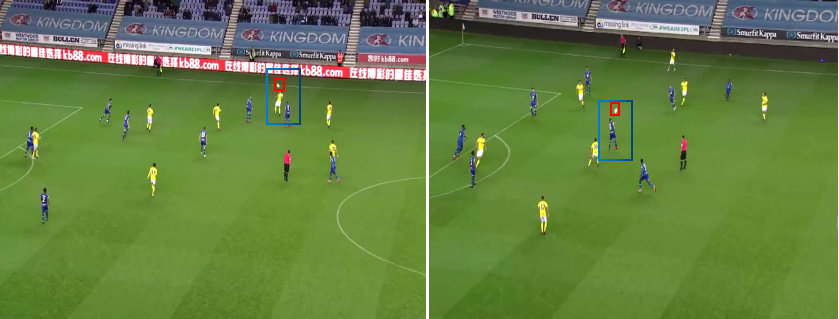}
\caption{Qualitative examples of header spotting. The red box indicates the detected ball, and the blue box highlights the player in the header action.}
\Description{Examples of header events with the detected ball and the player performing the header highlighted.}
\label{fig:qualitative_examples}
\end{figure}

\begin{figure}[t]
\centering
\includegraphics[width=\linewidth]{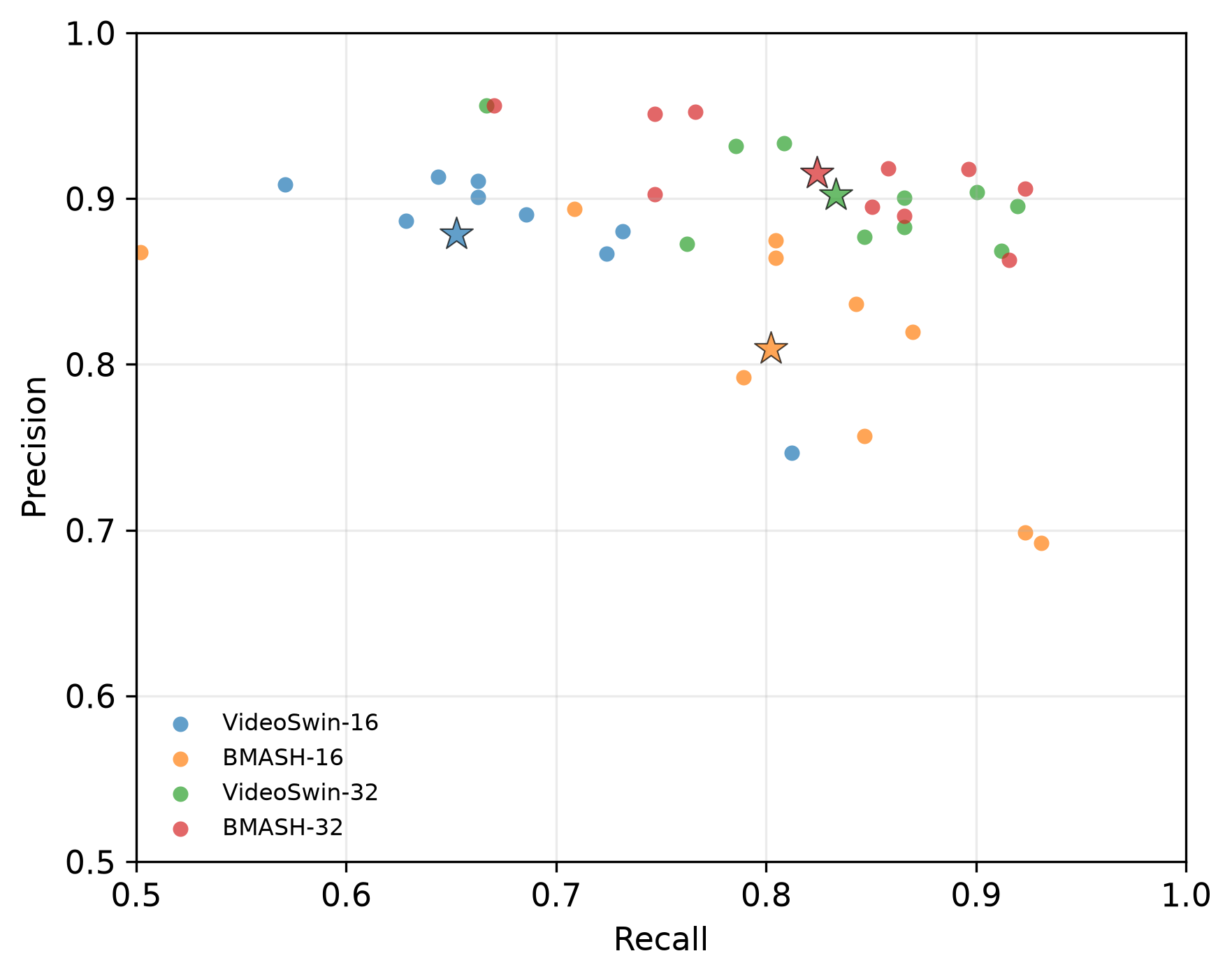}
\caption{Precision--recall results across folds. Points show individual folds and stars show the mean for each model.}
\Description{Precision and recall results across evaluation folds for the evaluated models.}
\label{fig:precision_recall_points}
\end{figure}

Figure~\ref{fig:per_fold_f1} shows the per-fold F1-score comparison between the proposed BMASH model and all evaluated baseline methods. Figure~\ref{fig:qualitative_examples} shows a representative header case, with the detected ball and header action highlighted. Figure~\ref{fig:precision_recall_points} shows the precision--recall across folds. BMASH-32 is concentrated in the upper-right region, indicating a stronger balance between precision and recall than the other evaluated models. This compact distribution across folds also indicates stable behavior under different game-level validation splits.

\subsection{Full-Video Header Spotting Analysis}
\label{sec:full_video_results}

In addition to temporal-window clip classification, we evaluate full-video header spotting on the two test broadcast matches. In this setting, the models process continuous video streams and convert dense sliding-window scores into discrete header-event predictions. The two videos contain a total of 261 annotated header events. 

Table~\ref{tab:full_video_pooled_results} summarizes the event-level results across both full broadcast videos. Video Swin-32 achieves the highest overall F1-score (0.636), together with the highest precision among the 32-frame models (0.617). BMASH-32 achieves almost the same F1-score (0.635) while detecting more true headers than Video Swin-32 (175 vs. 171) and obtaining the highest recall (0.671), at the cost of more false positives. Video Swin-16 attains the same recall as BMASH-32 but produces substantially more false positives, whereas BMASH-16 achieves the highest precision (0.621) and the fewest false positives (97), although with reduced recall (0.609). Overall, these results illustrate the trade-off between recall and precision in full-video header spotting.

\begin{table}[t]
\centering
\caption{Full-video header spotting results across the two test videos, using two second temporal matching.}
\label{tab:full_video_pooled_results}
\small
\scriptsize
\setlength{\tabcolsep}{6pt}
\begin{tabular}{lrrrrrrrr}
\toprule
Model & GT & Pred & TP & FP & FN & P & R & F1 \\
\midrule
BMASH-32 & 261 & 290 & \textbf{175} & 115 & \textbf{86} & 0.603 & \textbf{0.671} & 0.635 \\
Video Swin-32 & 261 & 277 & 171 & 106 & 90 & 0.617 & 0.655 & \textbf{0.636} \\
BMASH-16 & 261 & 256 & 159 & \textbf{97} & 102 & \textbf{0.621} & 0.609 & 0.615 \\
Video Swin-16 & 261 & 302 & 175 & 127 & \textbf{86} & 0.580 & 0.671 & 0.622 \\
\bottomrule
\end{tabular}
\end{table}

Compared with temporal-window classification, full-video spotting is substantially more challenging because dense sliding-window inference over long broadcast videos accumulates false positive detections and introduces temporal localization errors. The addition of ball-motion features produces performance comparable to the Video Swin baseline, with BMASH-32 favoring higher recall while Video Swin-32 provides a slightly better balance between precision and recall.

Table~\ref{tab:full_video_per_video_results} presents the results for each test video separately. Performance varies across the two matches, indicating that full-video spotting is influenced by factors such as camera viewpoint, event density, ball visibility, and visually similar aerial-ball situations. All models achieve higher precision on WB than on BN, while BN generally produces more false positive detections.

\begin{table}[t]
\centering
\caption{Full-match header spotting results on the two  broadcast test videos, reported separately for each video.}
\label{tab:full_video_per_video_results}
\scriptsize
\setlength{\tabcolsep}{5pt}

\begin{tabular}{llrrrrrrrr}
\toprule
Video & Model & GT & Pred & TP & FP & FN & P & R & F1 \\
\midrule

\multirow{4}{*}{BN}
& BMASH-32      & 111 & 133 & 71  & 62 & 40 & 0.534 & 0.640 & 0.582 \\
& BMASH-16      & 111 & 137 & 75  & 62 & 36 & 0.547 & 0.676 & 0.605 \\
& Video Swin-32 & 111 & 127 & 70  & \textbf{57} & 41 & \textbf{0.551} & 0.631 & 0.588 \\
& Video Swin-16 & 111 & 160 & \textbf{85}  & 75 & \textbf{26} & 0.531 & \textbf{0.766} & \textbf{0.627} \\

\midrule

\multirow{4}{*}{WB}
& BMASH-32      & 150 & 157 & \textbf{104} & 53 & \textbf{46} & 0.662 & \textbf{0.693} & \textbf{0.678} \\
& BMASH-16      & 150 & 119 & 84  & \textbf{35} & 66 & \textbf{0.706} & 0.560 & 0.625 \\
& Video Swin-32 & 150 & 150 & 101 & 49 & 49 & 0.673 & 0.673 & 0.673 \\
& Video Swin-16 & 150 & 142 & 90  & 52 & 60 & 0.634 & 0.600 & 0.616 \\

\bottomrule
\end{tabular}
\end{table}

\section{Conclusion}
\label{sec:conclusion}

We presented BMASH, a ball-motion-aware framework for soccer header spotting in broadcast videos. BMASH combines Video Swin-based action representations with detector-derived ball-motion descriptors using lightweight late fusion, allowing ball motion to support the Video Swin based action understanding. We evaluate BMASH against EfficientNet3D, TSM-ResNet50, and Video Swin baselines using game-level splits, with both temporal-window clip-level classification and full-video spotting analysis. 

At the clip windows level, BMASH-32 improved average precision from 0.9423 to 0.9530 and ROC-AUC from 0.9418 to 0.9507 compared with the matched Video Swin-32 baseline, while maintaining almost similar F1-score. This indicates that detector-derived ball features improve the ranking of header and non-header samples. In the full-video setting, Video Swin-32 and BMASH-32 achieved nearly similar event-level F1-scores, with BMASH-32 detecting more true headers and obtaining higher recall. These results highlight the difficulty of continuous full-match spotting, where overlapping windows, visually similar airborne-ball actions, missed ball detections, and many false positives affect event-level performance. 

The present study has some limitations. Full-video spotting was evaluated on only two held-out broadcast matches, so larger and more diverse full-match evaluations are needed to better assess generalization. In addition, the current ball descriptor summarizes simple detector statistics, such as visibility, position, scale, and displacement, but does not yet model full ball trajectories, speed, acceleration, or player-relative ball geometry. 

Future work will extend the training and evaluation to larger and more diverse datasets, investigate trajectory-aware and player-relative ball representations, improve post-processing for continuous spotting, and explore the integration of video predictions with wearable-sensor measurements for downstream head-impact analysis.

% sensor measurements
\begin{acks}

This work was part of Finland's Ministry of Education and Culture’s Doctoral Education Pilot under Decision No. VN/3137/2024-OKM-6 (The Finnish Doctoral Program Network in Artificial Intelligence, AI-DOC). We thank CSC – IT Center for Science, Finland, for providing the computational resources used in this work. We also thank Vili Pesonen, Ronja Suomala, and Miikka Pasanen for their help with the manual annotation and verification of soccer header events in the broadcast videos used in this study.
\end{acks}

\bibliographystyle{ACM-Reference-Format}

\bibliography{references}

\end{document}